\documentclass{bmvc2k}
\usepackage{times}
\usepackage{soul}
\usepackage{url}
\usepackage{graphicx}
\usepackage{amsmath}
\usepackage{amssymb,bm}
\usepackage{amsthm}
\usepackage{tabu}
\usepackage{booktabs}

\newcommand{\mR}{\mathcal{R}}
\newcommand{\mt}{\bm{t}}
\newcommand{\mP}{\bm{p}}
\newcommand{\mU}{\bm{u}}
\newcommand{\mV}{\bm{v}}
\newcommand{\mk}{\bm{k}}
\newcommand{\mK}{\bm{K}}
\newcommand{\mL}{\mathcal{L}}
\newcommand{\mI}{\mathcal{I}}
\newcommand{\mM}{\mathcal{M}}

\newcommand{\mf}{\bm{f}}

\title{\vspace{-0.15cm}6DoF Object Pose Estimation via Differentiable Proxy Voting Loss\vspace{-0.15cm}}

\addauthor{Xin Yu\vspace{-0.5cm}}{}{1}
\addauthor{Zheyu Zhuang\vspace{-0.5cm}}{}{2}
\addauthorr{Piotr Koniusz\vspace{-0.5cm}}{}{3}{2}
\addauthor{Hongdong Li\vspace{-0.5cm}}{}{2}

\addinstitution{
University of Technology Sydney
}
\addinstitution{
Australian National University}
\addinstitution{
Data61/CSIRO, Australia}

\runninghead{Xin Yu \etal}{6DoF Obj. Pose Est. via Differentiable Proxy Voting Loss}

\def\eg{\emph{e.g}\bmvaOneDot}

\def\etal{\emph{et al}\bmvaOneDot}
\def\ie{\emph{i.e}\bmvaOneDot}

\begin{document}

\maketitle

\begin{abstract}
Estimating a 6DOF object pose from a single image is very challenging due to occlusions or textureless appearances. 
Vector-field based keypoint voting has demonstrated its effectiveness and superiority on tackling those issues. 
However, direct regression of vector-fields neglects that the distances between pixels and keypoints also affect the deviations of hypotheses dramatically. In other words, small errors in direction vectors may generate severely deviated hypotheses when pixels are far away from a keypoint.
In this paper, we aim to reduce such errors by incorporating the distances between pixels and keypoints into our objective. 
To this end, we develop a simple yet effective differentiable proxy voting loss (DPVL) which mimics the hypothesis selection in the voting procedure.
By exploiting our voting loss, we are able to train our network in an end-to-end manner.
Experiments on widely used datasets, \ie, LINEMOD and Occlusion LINEMOD, manifest that our DPVL improves pose estimation performance significantly and speeds up the training convergence. 
\end{abstract}

\vspace{-0.3cm}
\section{Introduction}
\vspace{-0.5em}
Object pose estimation aims at obtaining objects' orientations and translations relative to a camera, and is widely used in many applications, such as robotic picking and virtual reality. Regarding an object may undergo severe occlusions, different lighting conditions or cluttered backgrounds, estimating its pose, including 3D orientations and translations, from an RGB image is quite difficult but has attracted great attention.

Conventional pose estimation methods exploit hand-crafted features~\cite{lowe2004distinctive} to establish the correspondences between an input image and a 3D model, and then estimate an object pose. However, hand-crafted features are often not robust to variations, such as different lighting conditions, cluttered background and occlusions. As a result, they may fail to estimate poses accurately or even localize objects in cluttered backgrounds. Thus, discriminative feature representations are highly desirable to account for those variations.

\begin{figure}[t]
\centering
\subfigure[]{\label{fig:openfig}\scalebox{0.72}[0.72]{\includegraphics[width=0.55\linewidth]{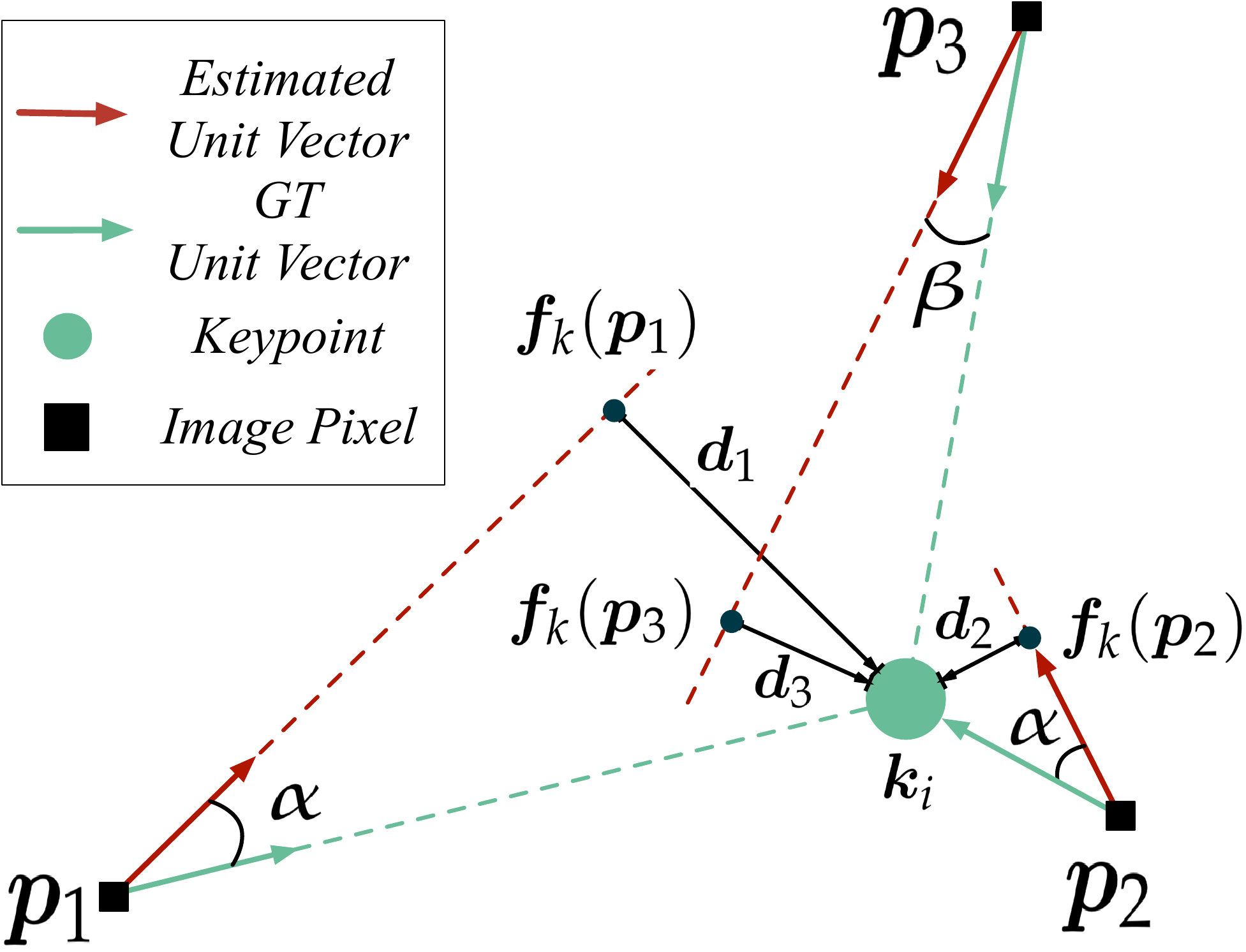}}}
\hspace{2.5em}
\subfigure[]{\label{fig:openfig2}\scalebox{0.72}[0.72]{\includegraphics[width=0.45\linewidth]{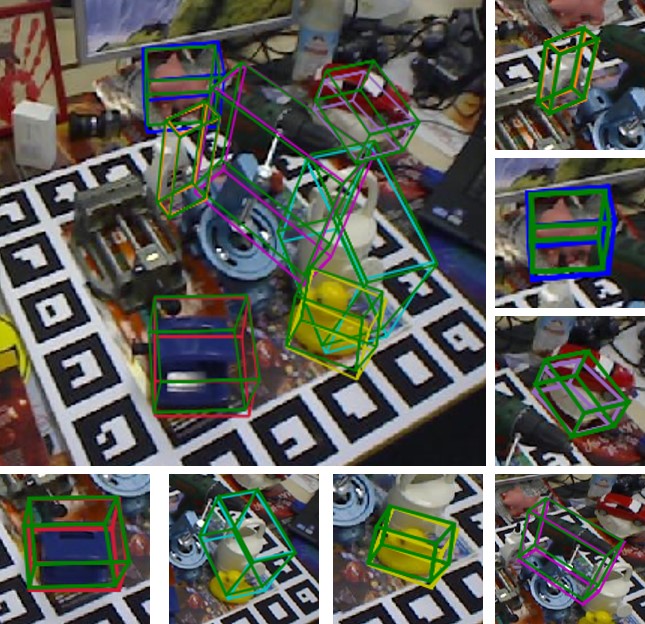}}}
\caption{Motivation of our differentiable proxy voting loss (a) and our results on the \emph{Occlusion LINEMOD} dataset (b). 
(a) when the estimation errors of direction vectors are the same (\eg, $\alpha$), the distance between a pixel and a keypoint affects the closeness between a hypothesis and the keypoint. (b) all the poses of the visible objects (except "Bench vise") are estimated. Green 3D bounding boxes indicate the ground-truth poses and those with other colors represent our predictions.}
\vspace{-0.3cm}
\end{figure}

Recently, deep learning based object pose estimation methods have been proposed recently~\cite{xiang2017posecnn,kehl2017ssd,peng2019pvnet}. Several methods leverage large training data to regress object poses or coordinates directly from input images~\cite{kehl2017ssd,tekin2018real,Park2019Pix2Pose,li2019CDPN,zakharovdpod}. Due to severe occlusions, the predicted poses or object coordinates may be inaccurate.
To increase the robustness to occlusions, voting strategies are employed to localize feature keypoints or coordinates~\cite{xiang2017posecnn,hu2019segmentation,peng2019pvnet}. 
Specifically, a keypoint is localized by a hypothesis with the highest vote from a vector field. In this manner, even invisible keypoints can be inferred from visible pixels. Considering some pixels are far away from a keypoint and their direction vectors exhibit small errors, previous methods may generate hypotheses (\ie, an intersection of the directional vectors of two pixels) with large deviations from the keypoint. As illustrated in Fig.~\ref{fig:openfig}, a smaller error $\beta$ leads to a larger deviation $d_3$ compared to $d_2$ resulted from a larger error $\alpha$.

In this paper, we propose a differentiable proxy voting loss (DPVL) to achieve accurate vector-field representations. DPVL takes the distances between pixels and keypoints into account to reduce hypothesis deviations caused by inaccurate direction vectors. Since sampling and inlier searching are involved in voting procedures, such as RANSAC~\cite{ransac}, they are often indifferentiable. Therefore, it is difficult to exploit the voted keypoints and their ground-truth counterparts as supervision to train networks in an end-to-end fashion. 

Instead of using voted keypoints in training, we develop a proxy hypothesis for each pixel to approximate the deviated distance between a voted keypoint (\ie, a hypothesis with the highest vote) and its ground-truth.
To be specific, we employ the foot of the perpendicular through a ground-truth keypoint with respect to a line generated by a pixel and its estimated direction vector as our proxy hypothesis. In this way, each pixel produces an approximated hypothesis with respect to a ground-truth keypoint. 
Note that, calculating the foot of the perpendicular for a keypoint is differentiable and the distance between a keypoint and a proxy hypothesis takes into account the estimated direction vector of a pixel as well as its distance to the keypoint. 
We therefore enforce a proxy hypothesis to be close to its corresponding keypoint in an end to end training manner. By doing so, we reduce the deviations of hypotheses and thus estimate keypoints accurately. 
Experiments on popular widely-used standard datasets also demonstrate that our DPVL not only improves the pose estimation performance significantly but also accelerates the training convergence.

\section{Related Work}
\vspace{-0.5em}
In this section, we mainly review single RGB image based 6DoF pose estimation approaches.

\noindent\textbf{Traditional methods: }
Conventional object pose estimation methods mainly leverage on the local feature/keypoint matching. The extracted local feature descriptors from 2D images, such as SIFT~\cite{lowe2004distinctive}, need to be robust to viewpoint, illumination, rotation and scale variations. After associating the extracted local features with points on a 3D model, the object poses are solved via a Perspective-n-Point (PnP) problem~\cite{lepetit2009epnp}.
However, these methods can only tackle textured objects, where local features are detectable. 
Other than local features, image templates~\cite{hinterstoisser2011multimodal,gu2010discriminative,rios2013discriminatively,zhu2014single}, and image edges~\cite{hinterstoisser2011gradient,liu2010fast} are also exploited for pose estimation.

\vspace{0.5em}
\noindent\textbf{Deep model based methods: }
Due to the powerful feature representation ability of deep neural networks, deep learning based methods have demonstrated impressive results on object pose estimation~\cite{kehl2017ssd,xiang2017posecnn,peng2019pvnet}.

Some methods, such as Viewpoints and Keypoints~\cite{tulsiani2015viewpoints} and Render for CNN~\cite{su2015render}, formulate the 3D pose estimation as a classification task by discretizing the pose space and then assigning objects with discrete pose labels. 
Motivated by the state-of-the-art image detection methods~\cite{liu2016ssd,ren2015faster}, pose estimation approaches are designed to localize objects while predicting their viewpoints based on the estimation of 3D bounding-boxes~\cite{kehl2017ssd,rad2017bb8,tekin2018real}, features of interest~\cite{xiang2017posecnn,peng2019pvnet} or coordinate maps~\cite{zakharovdpod,wang2019normalized,Park2019Pix2Pose}.
SSD6D~\cite{kehl2017ssd} extends the ideas of 2D object detection and classifies localized objects with discrete poses while YOLO6D~\cite{tekin2018real} regresses 3D bounding-boxes of objects. 
BB8~\cite{rad2017bb8} firstly generates 2D segmentation masks for objects and predicts 3D bounding-boxes from the 2D masks. CPDN~\cite{li2019CDPN}, DPOD~\cite{zakharovdpod} and Pix2Pose~\cite{Park2019Pix2Pose} regress the 2D/3D coordinates of 3D object models from images. Object poses are estimated by PnP after obtaining 2D-3D correspondences.

Regarding directly regressing the 2D projections of 3D points is not reliable, PoseCNN~\cite{xiang2017posecnn} firstly estimates a vector-field pointing to an object center from object pixels and then employs Hough voting to determine the center. The translations and rotations of the object are regressed subsequently by a subnetwork. Rather than only estimating a centroid, PVNet~\cite{peng2019pvnet} votes several features of interest, while the work~\cite{hu2019segmentation} votes the corners of a 3D bounding-box from each segmentation grid. Due to the voting strategy, the locations of estimated 2D projections are more robust to occlusions. However, small errors in a direction vector field may lead to large deviations of hypotheses.

\vspace{0.5em}
\noindent\textbf{Pose refinement: }
Depth images are used to refine estimated poses. For instance, PoseCNN employs the depth information and Iterative Closest Point (ICP)~\cite{besl1992method} to refine the estimated poses. 
Meanwhile, DeepIM~\cite{li2018deepim} and DPOD~\cite{zakharovdpod} exploit optical flow to modify initial pose estimation by minimizing the differences between the object appearance and the 2D projection of the 3D model. Moreover, Hu~\etal~\cite{hu2019single} develop a standalone pose refinement network by taking initial pose estimation from other methods. Hybrid-pose~\cite{song2020hybridpose} improves PVNet by incorporating different pose representations and an additional refinement sub-network.
Note that, those methods are effective when initial poses are close to their ground-truths. Therefore, generating accurate initial poses also plays a key role in the pose refinement pipeline.

\section{Proposed Method}
\vspace{-0.5em}
In this paper, we focus on obtaining accurate initial pose estimation. In particular, our method is designed to localize and estimate the orientations and translations of an object accurately without any refinement. The object pose is represented by a rigid transformation $(\mR, \mt)$ from the object coordinate system to the camera coordinate system, where $\mR$ and $\mt$ indicate 3D rotations and translations, respectively.

\begin{figure*}[t]
\centering
\scalebox{0.95}[0.95]{\includegraphics[width=1\linewidth]{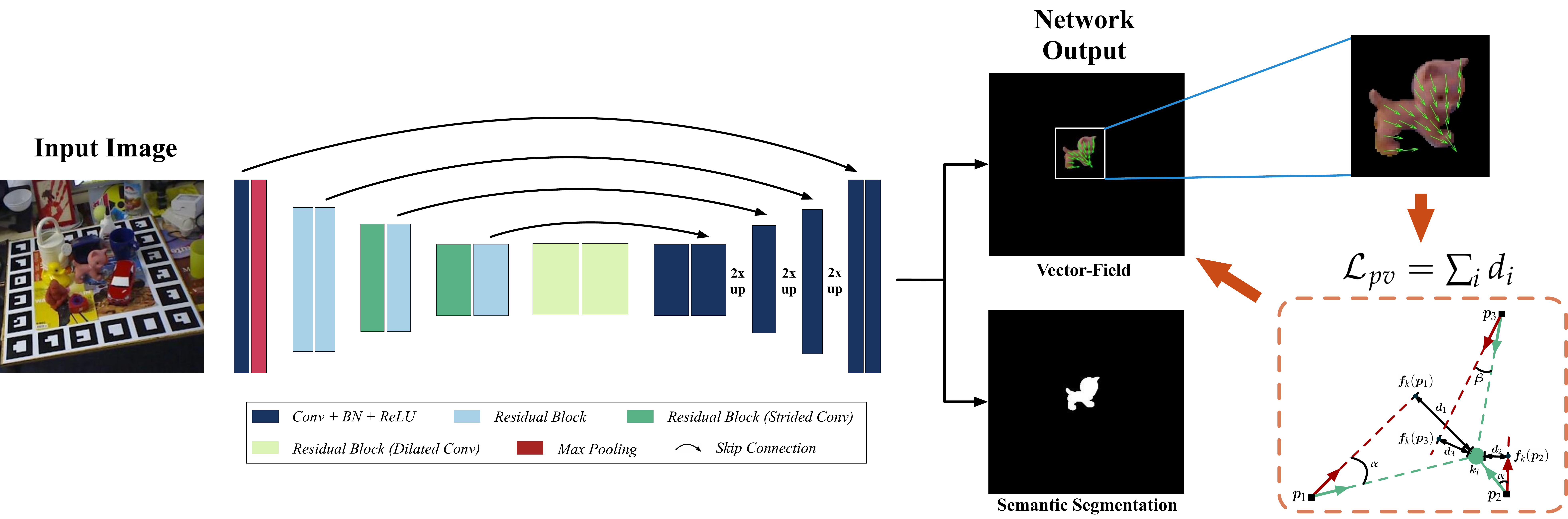}}
\vspace{-0.8em}
\caption{Illustration of our pipeline. Here, we only highlight our proposed differentiable proxy voting loss (DPVL). For simplicity, the vector-field regression loss and segmentation loss are omitted.}
\vspace{-1em}
\label{fig:pipeline}
\end{figure*}

Since voting based methods~\cite{xiang2017posecnn,peng2019pvnet} have demonstrated their robustness to occlusions and view changes, we therefore follow the voting based pose estimation pipeline. Specifically, our method firstly votes 2D positions of the object keypoints from the vector-fields (as illustrated by the green arrows in Fig.~\ref{fig:pipeline}) and then estimates the 6DOF pose by solving a PnP problem. Prior works regress pixel-wise vector-fields by an $\ell_1$ loss. However, small errors in the vector-fields may lead to large deviation errors of hypotheses because the loss does not take the distance between a pixel and a keypoint into account. Therefore, we present a differentiable proxy voting loss (DPVL) to reduce such errors by mimicking the hypothesis selection in the voting procedure. Furthermore, benefiting from our DPVL, our network is able to converge much faster.

\vspace{-0.5em}
\subsection{Vector-fields of keypoints}
\vspace{-0.5em}
In order to estimate object poses, we firstly localize 2D keypoints of a given object. Regarding the sizes of objects vary significantly, we opt to estimate a vector-field for each keypoint instead of predicting their keypoint positions from images. The vector-field is composed of unit direction vectors pointing from a pixel to a certain keypoint. Different from feature keypoints extracted from image local patches, our direction vectors are estimated by a deep neural network with a larger receptive field covering a large portion of objects. In this manner, even though a keypoint is not visible, it is able to be deduced from the visible parts of an object.

Motivated by the works \cite{xiang2017posecnn,peng2019pvnet}, our network simultaneously outputs a segmentation mask for an object and keypoint vector-fields, as shown in Fig.~\ref{fig:pipeline}.  
Specifically, if a pixel $\mP$ belongs to an object, its segmentation mask will be assigned to 1. Otherwise, its segmentation mask is assigned to 0. A unit direction vector $\mU_{k}(\mP)$ pointing from a pixel $\mP$ to a keypoint $\mk$ is expressed as $\mU_{k}(\mP) = (\mk - \mP)/{\|\mk - \mP\|_2}$.

\vspace{-0.5em}
\subsection{Differentiable proxy voting loss}
\vspace{-0.5em}
Given an image $\mI$ and the keypoint locations of an object $\mK=\{\mk_i\}, i=1,\cdots,N$,\footnote{For notation convenience, we will drop the subscript $i$ in $\mk_i$.} where $N$ represents the number of the chosen keypoints, the ground-truth keypoint vector-fields are easily derived while the estimated vector-fields are generated from our network. Then, the smooth $\ell_1$ loss \cite{girshick2015fast} is employed to regress the ground-truth direction vectors as:
\begin{equation}
\begin{split}
\!\!\mL_{vf} & = \sum_{\mk\in \mK}\sum_{\mP\in\mM} \ell_1(\|\mU_{k}(\mP) - \mV_{k}(\mP)\|_1),\\
\!\!\ell_1({a}) & =  0.5a^2\mathbb{I}(|a|\!<\!1)\!+\! (|a|\!-\!0.5)\mathbb{I}(|a|\!\geq \!1),
\end{split}
\end{equation}
where $\mV_k(\mP)$ represents an estimated direction vector, $\mM$ indicates an object mask, $\mathbb{I}(\cdot)$ is an indicator function and $a$ represents a scalar variable.

As aforementioned, small errors in the estimated unit direction vectors may lead to large deviations of hypotheses (as illustrated by $d_2$ and $d_3$ in Fig.~\ref{fig:openfig}), and diverse hypotheses will result in inaccurate keypoints, thus decreasing the performance of pose estimation. 
Unlike previous works, we take the distributions of hypotheses into account and enforce all the hypotheses to be close to the ground-truth keypoints. Assuming an object contains $M$ pixels, there will be $M(M-1)$ hypotheses. Although there is a closed-form solution to obtaining a hypothesis given two direction vectors from two pixels, calculating all the hypotheses leads to inefficient training of deep networks especially when the resolution of an object is very large. 

Since the distance between a keypoint and a point on a line is not upper-bounded but lower-bounded, we opt to use the foot of the perpendicular through a keypoint with respect to a line specified from a direction vector of a pixel $\mP$ to approximate a hypothesis. 
Note that, the distance $d$ (in Fig.~\ref{fig:openfig}) is the lower bound distance between a keypoint and a line.
By doing so, we only need to compute $M$ perpendiculars rather than $M(M-1)$ hypotheses, significantly reducing computations. 
More importantly, the distance between a keypoint $\mk$ and its corresponding foot of the perpendicular $\mf_k(\mP)$ with respect to the direction vector $\mV_{k}(\mP)$ of pixel $\mP$ has a close-form solution and is also differentiable. Therefore, we minimize the distance, as our differentiable proxy voting loss (DPVL) $\mL_{pv}$, to force the proxy hypotheses to be close to the keypoint as follows:
\vspace{-0.3cm}
\begin{equation}
\label{eqn1}
\mL_{pv} = \sum_{\mk\in \mK}\sum_{\mP\in \mM}\ell_1(\|\mk - \mf_k(\mP)\|_1) = \sum_{\mk\in \mK}\sum_{\mP\in \mM}\ell_1\left(\frac{|v_k^y k^x-v_k^x k^y+v_k^x p^y - v_k^y p^x |}{\sqrt{(v_k^x)^2 + (v_k^y)^2)}}\right),
\end{equation}
where $\mV_k(\mP) = (v_k^x, v_k^y)$, $\mP = (p^x, p^y)$, $\mk = (k^x, k^y)$ and the superscripts $x, y$ represent horizontal and vertical coordinates respectively. 
Since $\mV_k(\mP)$ is directly estimated from our network and it might not be a unit vector, there is a normalization operation in Eqn.~\eqref{eqn1} (The detailed derivations of Eqn.~\eqref{eqn1} are provided in the supplementary material).
As indicated by Eqn.~\eqref{eqn1}, our developed loss $\mL_{pv}$ not only enforces the direction vector $\mV_k(p)$ to point to keypoints but also is sensitive to the pixel locations. In other words, small errors in the direction vectors of pixels away from a keypoint will be penalized more severely in order to produce more concentrated hypotheses. 

\subsection{Network architecture and training strategy}
To demonstrate the effectiveness of our proposed loss, we adopt the same architecture as PVNet~\cite{peng2019pvnet}, as illustrated in Fig.~\ref{fig:pipeline}. The network is built on a pre-trained ResNet18 backbone~\cite{he2016deep} with three subsequent upsampling layers, and outputs segmentation predictions $s(\mP)\in[0,1]$ and the vector-fields of keypoints. Our overall objective is written as:
\begin{equation}
\label{eqn2}
\mL = \mL_{seg} + \mL_{vf} + \lambda\mL_{pv},
\end{equation}
where $\mL_{seg} = -\sum_{\mP\in\mM}\log(s(\mP))$ is the segmentation loss, and $\lambda$ is a trade-off weight. Following the training protocols of PVNet, we set the same weights to $\mL_{seg}$ and $\mL_{pv}$. In our experiments, we set $\lambda$ to $1e^{-3}$. Since the objective is a multi-task problem, the gradients of these two tasks are balanced using gradient normalization~\cite{chen2017gradnorm}.
Note that, although we only use one more term (\ie, DPVL) in Eqn.~\eqref{eqn2} compared to PVNet while using the same network architecture, it makes the training efficiency and final performance significantly different. In particular, it speeds up the convergence of training and improves estimation accuracy of vector fields without introducing any other network parameters. 

\vspace{-1em}
\subsection{DPVL as a regularizer}
Since our proposed DPVL is designed to obtain generated hypothesis with small deviations, it is not directly to force the estimated vector-fields to be close to the ground-truth ones. Therefore, we remove the regression loss $\mL_{vf}$ in Eqn.~\ref{eqn2} and examine whether DPVL can be used as a standalone loss. We found that by only using DPVL to estimate vector-fields, our network fail to converge. 
Note that, a direction vector and its reversed one will result in the same loss. Without employing the vector-field regression, our network will suffer this ambiguity and fail to produce consistent vector-fields when minimizing the DPVL. In other words, DPVL cannot be used as a standalone loss and it requires the vector-field regression loss to reduce the ambiguity.
Therefore, our proposed DPVL is employed as a regularization term for accurate vector-field estimation.

\vspace{-1em}
\subsection{Implementation details}
For a fair comparison with our baseline method PVNet, we select 8 keypoints for each object by running the farthest point sampling algorithm on its 3D model. Following \cite{peng2019pvnet}, we render 10,000 images and synthesize 10,000 images by ``Cut and Paste" for each object. Data augmentation, including random cropping, resizing, rotation and color jittering, is employed to prevent overfitting. 

In training, we set the batch size to 16 and employ Adam optimizer with an initial learning rate $1e^{-3}$. The learning rate is decayed to $1e^{-5}$ by a factor of $0.85$ every 5 epochs. Our maximum training epoch is set to $100$, whereas PVNet requires $200$ epochs.
In testing, Hough voting is employed to localize keypoints and then EPnP~\cite{lepetit2009epnp} is used to solve 6DOF poses. 
Similar to PVNet, our method also runs in real-time on a test image of 480$\times$640 pixels on a GTX 2080 Ti GPU.

\section{Experiments}
\vspace{-0.5em}
We conduct experiments on two popular widely used datasets that are designed to evaluate 6DOF pose estimation methods.

\vspace{-0.5em}
\subsection{Datasets}
\noindent{\bf LINEMOD}~\cite{hinterstoisser2012model} is a \emph{de facto} 6DOF pose estimation benchmark consisting of 15783 images for 13 objects (around 1,200 instances for each object). CAD models of those objects are also provided. LINEMOD exhibits many challenges including cluttered scenes, textureless objects, and lighting variations.

\noindent{\bf Occlusion LINEMOD}~\cite{brachmann2014learning} provides additional annotations on a subset of LINEMOD images. Each image contains multiple annotated objects and those objects are severely occluded due to the scene clutter, posing extreme challenges for pose estimation.

\begin{table*}[t]\renewcommand{\arraystretch}{0.9}
\begin{center}
\caption{Pose estimation performance on the \emph{LINEMOD} dataset w.r.t. ADD(-S) scores. Glue and egg box are considered as symmetric.}
{\scriptsize
\begin{tabular}{>{\centering\arraybackslash}m{0.12\linewidth}|
>{\centering\arraybackslash}m{0.05\linewidth}|
>{\centering\arraybackslash}m{0.05\linewidth}|
>{\centering\arraybackslash}m{0.05\linewidth}|
>{\centering\arraybackslash}m{0.05\linewidth}|
>{\centering\arraybackslash}m{0.05\linewidth}|
>{\centering\arraybackslash}m{0.05\linewidth}|
>{\centering\arraybackslash}m{0.05\linewidth}|
>{\centering\arraybackslash}m{0.05\linewidth}|
>{\centering\arraybackslash}m{0.05\linewidth}}
\toprule
Methods & \scriptsize{BB8} & \!\!\scriptsize{SSD6D} &\!\!\!\!\scriptsize{YOLO6D} & \scriptsize{DPOD} &\!\!\!\scriptsize{Pix2Pose} & \scriptsize{CDPN} &\!\!\!\!\scriptsize{PoseCNN} & \scriptsize{PVNet} &  \scriptsize{Ours} \\
\midrule
\scriptsize{Ape}       &  40.4 & 65 & 21.62 & 53.28 & 58.1 & 64.38 & 27.8 & 43.62 & {\bf 69.05}\\ 
\scriptsize{Bench vise} &  91.8 & 80 & 81.80 & 95.34 & 91.0 & 97.77 & 68.9 & 99.90 & {\bf 100.0}\\
\scriptsize{Cam}       &  55.7 & 78 & 36.57 & 90.36 & 60.9 & 91.67 & 47.5 & 86.86 & {\bf 94.12}\\
\scriptsize{Can}       &  64.1 & 86 & 68.80 & 94.10 & 84.4 & 95.87 & 71.4 & 95.47 & {\bf 98.52}\\
\scriptsize{Cat}       &  62.6 & 70 & 41.82 & 60.38 & 65.0 & {\bf 83.83} & 56.7 & 79.34 & 83.13\\
\scriptsize{Driller}   &  74.7 & 73 & 63.51 & 97.72 & 73.6 & 96.23 & 65.4 & 96.43 & {\bf 99.01}\\
\scriptsize{Duck}      &  44.3 & 66 & 27.23 & 66.01 & 43.8 & {\bf 66.76} & 42.8 & 52.58 & 63.47\\
\scriptsize{Egg box}    &  57.8 & {\bf 100} & 69.58 & 99.72 & 96.8 & 99.72 & 98.3 & 99.15 & {\bf 100.0}\\
\scriptsize{Glue}      &  41.2 & {\bf 100} & 80.02 & 93.83 & 79.4 & 99.61 & 95.6 & 95.66 & 97.97\\
\scriptsize{Hole puncher}& 67.2 & 49 & 42.63 & 65.83 & 74.8 & 85.82 & 50.9 & 81.92 & {\bf 88.20}\\
\scriptsize{Iron}      &  84.7 & 78 & 74.97 & 99.80 & 83.4 & 97.85 & 65.6 & 98.88 & {\bf 99.90}\\
\scriptsize{Lamp}      &  76.5 & 73 & 71.11 & 88.11 & 82.0 & 97.86 & 70.3 & 99.33 & {\bf 99.81}\\
\scriptsize{Phone}     &  54.0 & 79 & 47.74 & 74.24 & 45.0 & 90.75 & 54.6 & 92.41 & {\bf 96.35}\\
\midrule
Mean      &  62.7 & 79 & 55.95 & 82.98 & 72.4 & 89.86 & 62.7 & 86.27 & {\bf 91.50}\\
\bottomrule
\end{tabular}
\label{tab1}
}
\end{center}
\vspace{-1.0em}
\end{table*}

\begin{figure*}[t]
\centering
{\includegraphics[width=1\linewidth]{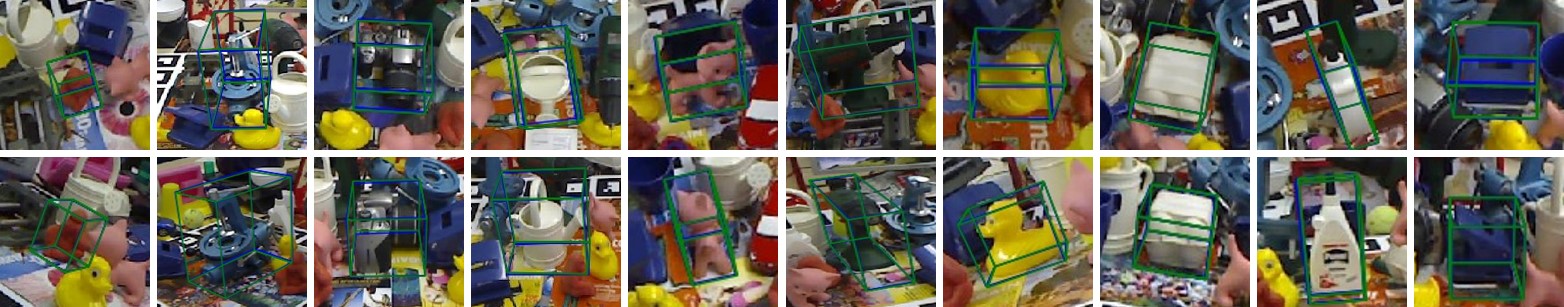}}
\vspace{-2em}
\caption{Visualization of our qualitative results on the \emph{LINEMOD} dataset. Green 3D bounding-boxes indicate the ground-truth poses and the blue ones represent our prediction.}
\label{fig:exp1}
\vspace{-1em}
\end{figure*}

\begin{table}[t]\renewcommand{\arraystretch}{0.9}
\centering
\begin{minipage}{0.62\textwidth}
\caption{Pose estimation performance on the \emph{LINEMOD} dataset w.r.t. 2D projection errors.}
{\scriptsize
\begin{tabular}{>{\centering\arraybackslash}m{0.15\linewidth}|
>{\centering\arraybackslash}m{0.07\linewidth}|
>{\centering\arraybackslash}m{0.08\linewidth}|
>{\centering\arraybackslash}m{0.07\linewidth}|
>{\centering\arraybackslash}m{0.08\linewidth}|
>{\centering\arraybackslash}m{0.07\linewidth}|
>{\centering\arraybackslash}m{0.07\linewidth}}
\toprule
\footnotesize{Methods} & \scriptsize{BB8} &\!\!\!\!\scriptsize{YOLO6D} & \scriptsize{CDPN} &\!\!\!\!\scriptsize{PoseCNN} & \scriptsize{PVNet} & \scriptsize{Ours} \\
\midrule
\scriptsize{Ape}       &  96.6 & 92.10 & 96.86 & 83.0 & 99.23 & {\bf 99.04}\\ 
\scriptsize{Bench vise} &  90.1 & 95.06 & 98.35 & 50.0 & {\bf 99.81} & 99.71\\
\scriptsize{Cam}       &  86.0 & 93.24 & 98.73 & 71.9 & 99.21 & {\bf 99.41}\\
\scriptsize{Can}       &  91.2 & 97.44 & 99.41 & 69.8 & 99.90 & {\bf 100.0}\\
\scriptsize{Cat}       &  98.8 & 97.41 & {\bf 99.80} & 92.0 & 99.30 & 99.70\\
\scriptsize{Driller}   &  80.9 & 79.41 & 95.34 & 43.6 & 96.92 & {\bf 98.12}\\
\scriptsize{Duck}      &  92.2 & 94.65 & 98.59 & 91.8 & 98.02 & {\bf 99.06}\\
\scriptsize{Egg box}    &  91.0 & 90.33 & 98.97 & 91.1 & 99.34 & {\bf 99.43}\\
\scriptsize{Glue}      &  92.3 & 96.53 & 99.23 & 88.0 & 98.45 & {\bf 99.51}\\
\scriptsize{Holepuncher}& 95.3 & 92.86 & 99.71 & 82.1 & {\bf 100.0} & {\bf 100.0}\\
\scriptsize{Iron}      &  84.8 & 82.94 & 97.24 & 41.8 & 99.18 & {\bf 99.69}\\
\scriptsize{Lamp}      &  75.8 & 76.87 & 95.49 & 48.4 & 98.27 & {\bf 99.14}\\
\scriptsize{Phone}     &  85.3 & 86.07 & 97.64 & 58.8 & {\bf 99.42} & {\bf 99.42}\\
\midrule
\footnotesize{Mean} &  89.3 & 90.37 & 98.10 & 70.2 & 99.00 & {\bf 99.40}\\
\bottomrule
\end{tabular}
\label{tab2}
}
\end{minipage}
\begin{minipage}{0.35\textwidth}
\caption{Impact of different $\lambda$ on the \emph{LINEMOD} dataset. }
\label{ablationstudy}
{\scriptsize
\begin{tabular}{>{\centering\arraybackslash}m{0.25\linewidth}|
>{\centering\arraybackslash}m{0.08\linewidth}|
>{\centering\arraybackslash}m{0.08\linewidth}|
>{\centering\arraybackslash}m{0.08\linewidth}|
>{\centering\arraybackslash}m{0.08\linewidth}}
\toprule
{Mean Value} & 0 & $1e^{-4}$ & $1e^{-3}$ & $1e^{-2}$ \\
\midrule
\centering
ADD(-S)   & \!\!86.27 & \!\!87.63 & \!\!91.50 & \!\!90.92 \\
Proj. err & \!\!99.00 & \!\!99.24 & \!\!99.40 & \!\!99.36 \\
\bottomrule

\end{tabular}
}
\end{minipage}
\end{table}

\begin{table}[t]\renewcommand{\arraystretch}{0.9}
\begin{center}
\caption{Pose estimation performance on the \emph{Occlusion LINEMOD} dataset w.r.t. ADD(-S) scores.}
{\scriptsize
\begin{tabular}{>{\centering\arraybackslash}m{0.11\linewidth}|
>{\centering\arraybackslash}m{0.07\linewidth}|
>{\centering\arraybackslash}m{0.07\linewidth}|
>{\centering\arraybackslash}m{0.07\linewidth}|
>{\centering\arraybackslash}m{0.07\linewidth}|
>{\centering\arraybackslash}m{0.07\linewidth}|
>{\centering\arraybackslash}m{0.07\linewidth}|
>{\centering\arraybackslash}m{0.07\linewidth}}
\toprule
\!\!\footnotesize{Methods} &\!\!\!\!\scriptsize{YOLO6D} &\!\!\scriptsize{PoseCNN} &\!\!\scriptsize{Oberweger} &\!\!\scriptsize{Pix2Pose} & \!\!\scriptsize{DPOD} & \!\!\scriptsize{PVNet}  & \scriptsize{Ours} \\
\midrule
\scriptsize{Ape}       &  2.48 & 9.6   & 17.6 &  {\bf 22.0} &-& 15.81 & 19.23\\ 
\scriptsize{Can}       &  17.48 & 45.2 & 53.9 & 44.7 &-& 63.30 & {\bf 69.76}\\
\scriptsize{Cat}       &  0.67 & 0.93  & 3.31 & {\bf 22.7} &-& 16.68 &  21.06\\
\scriptsize{Driller}   &  7.66 & 41.4  & 62.4 & 44.7 &-& 65.65 & {\bf 71.58}\\
\scriptsize{Duck}      &  1.14 & 19.6  & 19.2 & 15.0 &-& 25.24 & {\bf 34.27}\\
\scriptsize{Egg box}    & -    &  22    & 25.9 & 25.2 &-& {\bf 50.17} & 47.32\\
\scriptsize{Glue}      &  10.08 & 38.5 & 39.6 & 32.4 &-& {\bf 49.62} & 39.65\\
\scriptsize{Hole puncher}& 5.45 & 22.1  & 21.3 & {\bf 49.5} & -& 39.67 &  45.27\\
\midrule
\footnotesize{Mean} &  6.42 & 24.9 & 30.4 & 32.0 & \!\!32.79 & 40.77 & {\bf 43.52}\\
\bottomrule
\end{tabular}
\label{tab3}
}
\end{center}
\vspace{-1.5em}
\end{table}

\begin{figure}[t]
\begin{minipage}{0.5\textwidth}
\centering
{\includegraphics[width=0.95\linewidth]{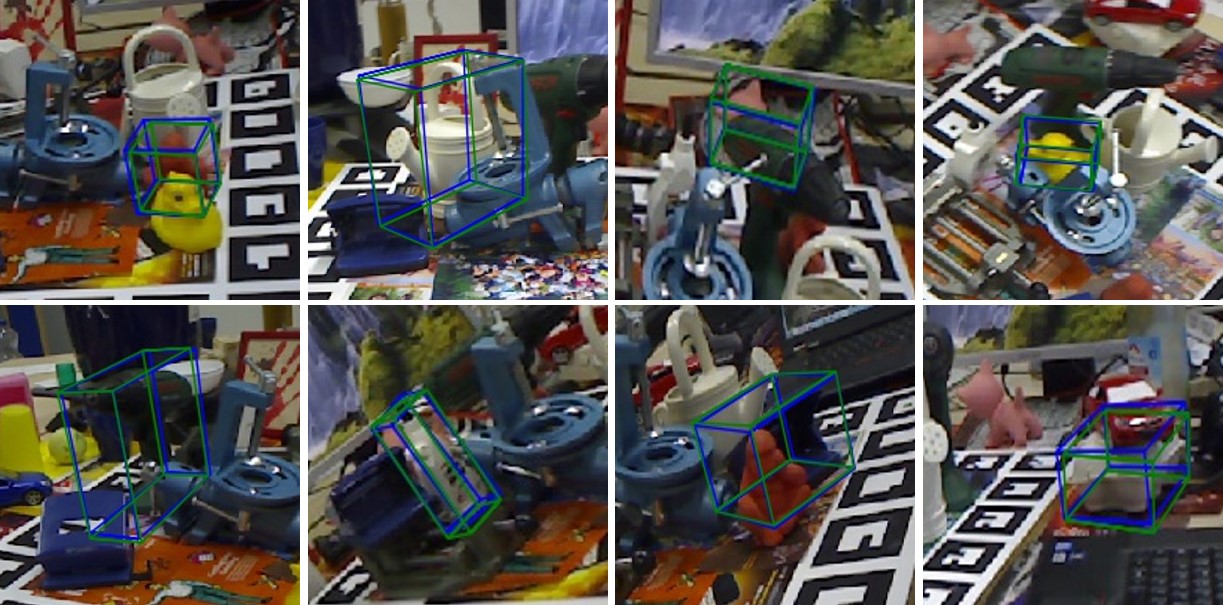}}
\vspace{-0.5em}
\caption{Our visual results on the \emph{Occlusion LINEMOD} dataset. Green 3D bounding-boxes indicate the ground-truth poses and blue ones are our predictions.\label{fig:exp3}}
\end{minipage}
\begin{minipage}{0.49\textwidth}
\subfigure[]{\scalebox{0.97}[1.2]{\includegraphics[width=0.5\linewidth, trim={2mm, 5mm, 3mm, 0mm}, clip]{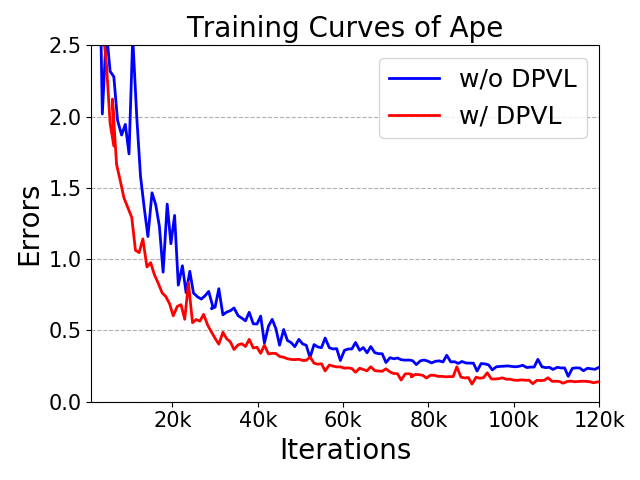}}}
\subfigure[]{\scalebox{0.97}[1.2]{\includegraphics[width=0.5\linewidth, trim={2mm, 5mm, 2mm, 0mm}, clip]{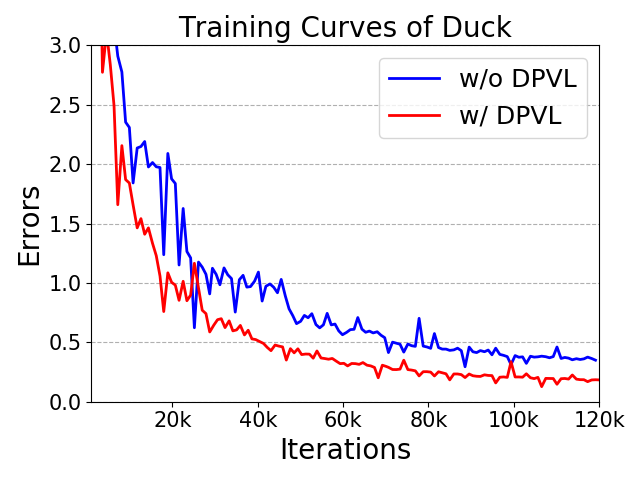}}}
\caption{Comparisons of the training curves of two objects with and without using our DPVL. The errors are measured by $\mL_{pv}$.\label{fig:training}}
\end{minipage}
\vspace{-1em}
\end{figure}

\vspace{-0.5em}
\subsection{Evaluation metrics}
We evaluate the performance of 6DOF estimation by using two commonly used metrics: {\bf 2D projection error} and {\bf ADD score}~\cite{hinterstoisser2012model}.
2D projection error measures the mean distance between the projections of a 3D model using the ground-truth and estimated poses. If the mean distance is less than 5 pixels, an estimated pose is regarded as correct. Thus, 2D projection error indicates the closeness between an object in the image and the projection of its 3D model using the estimated pose.
ADD score measures average 3D distance between 3D model points transformed by the ground-truth pose and the estimated one. When the distance is less than 10 percent of the diameter of the 3D model, the estimated pose is considered as correct. 
For a symmetric object, ADD(-S) is employed, where the 3D distance is calculated based on the closest point distance. 

\vspace{-0.5em}
\subsection{Comparisons with the state-of-the-art}
Since our method focuses on providing an accurate initial pose estimation from an RGB image, we do not adopt any pose refinement methods to further improve the estimation performance. Furthermore, as mentioned in the work~\cite{li2018deepim}, the performance of pose refinement also highly relies on good initial pose estimation. Thus, below we compare with the state-of-the-art RGB image based pose estimation methods without using any pose refinements.

\vspace{0.2em}
\noindent{{\bf Comparisons on LINEMOD: }}
We compare our algorithm with BB8~\cite{rad2017bb8}, SSD6D~\cite{kehl2017ssd}, YOLO6D~\cite{tekin2018real}, DPOD~\cite{zakharovdpod}, Pix2Pose~\cite{Park2019Pix2Pose}, CDPN~\cite{li2019CDPN}, PoseCNN~\cite{xiang2017posecnn}, and PVNet~\cite{peng2019pvnet} on ADD(-S) scores (in Table~\ref{tab1}) and 2D projection errors (in Table~\ref{tab2}). BB8, SSD6D and YOLO6D localize keypoints by regression, while DPOD, Pix2Pose and CDPN regress 3D object coordinates. PoseCNN firstly localizes object centers via Hough voting and then estimates poses by regression.  
PVNet and our method employ voting mechanisms to localize all the keypoints for pose estimation. 

Note that, BB8 and SS6D require necessary refinements to obtain satisfying results. Thus, we report their results with pose refinements. Since some methods do not report their 2D projection errors, we do not include those in Table~\ref{tab2}.
As seen in Table~\ref{tab1} and Table~\ref{tab2}, our method outperforms the state-of-the-art methods on both of the metrics. Figure~\ref{fig:exp1} illustrates our qualitative results (more qualitative results are shown in the supplementary material).

More importantly, our method achieves better performance on all the objects in terms of ADD(-S) scores compared to our baseline method PVNet, and improves the performance by a large margin of 5.23\%.
Our method also outperforms PVNet in terms of 2D projection errors except one class. 
As indicated in Table~\ref{tab1}, it is difficult to estimate accurate vector-fields from textureless objects, such as "ape" and "duck", by PVNet. On the contrary, our method is able to estimate vector-fields more accurately and thus obtains better pose estimation performance than PVNet.
In particular, our method improves the accuracy by {\bf 25.43\%} and {\bf 10.89\%} on "ape" and "duck" respectively in terms of ADD(-S) scores.

\vspace{0.2em}
\noindent{{\bf Comparisons on Occlusion LINEMOD: }}
Following previous methods (\ie, YOLO6D, PoseCNN, Oberweger\cite{oberweger2018making}, Pix2Pose, DPOD and PVNet), we directly apply the models trained on the LINEMOD dataset to the Occlusion LINEMOD dataset. Note that, we do not re-train our network on occluded data. Thus, for fair comparisons, we do not compare with Hybrid-pose~\cite{song2020hybridpose} since it is trained on occluded data and uses pose refinements.
ADD(-S) scores are often reported on this dataset and the comparisons of the state-of-the-art approaches are shown in Table~\ref{tab3}. Our method achieves the best overall performance compared to the state-of-the-art. 
In Fig~\ref{fig:openfig2}, we demonstrate the estimate results of all the objects (except "Bench vise") from one image.
Figure~\ref{fig:exp3} demonstrates our qualitative results on the Occlusion LINEMOD dataset (more visual results are shown in the supplementary material).
As expected, our method outperforms PVNet, the second best, thus demonstrating the effectiveness of our proposed DPVL.
Note that, due to the heavy occlusion, inaccurate segmentation (\ie, object localization) is an important reason accounting for erroneous pose estimation. Improving the segmentation performance is thus left for our future work.

\vspace{-0.5em}
\subsection{Ablation study}

By incorporating our proposed DPVL, our method is able to attain more accurate pose estimation and improve the training convergence. 
Since our network architecture and training data (the synthetic and rendered images are generated by the code provided by the authors of PVNet) are as the same as PVNet, the significant performance improvements mainly benefit from our proposed loss, as indicated in Table~\ref{tab1}, Table~\ref{tab2} and Table~\ref{tab3}.

We also conduct an ablation study on the weight $\lambda$. Here, we set the weight $\lambda$ to $0, 1e^{-2}, 1e^{-3}$ and $1e^{-4}$ to demonstrate how DPVL affects the network performance in Table~\ref{ablationstudy}. Note that, when $\lambda$ is $0$, the training objective of the network is identical to PVNet. When $\lambda$ is large, \ie, $1e^{-2}$, our DPVL $\mL_{pv}$ may overwhelm the vector field regression term $\mL_{vf}$ and lead to inferior vector field estimation. However, using too small $\lambda$, \ie, $1e^{-4}$, the impact of our DPVL is still marginal. Therefore, we set $\lambda$ to $1e^{-3}$.

In Fig.~\ref{fig:training}, we demonstrate that by using our DPVL, our method obtains lower distance errors between keypoints and proxy hypotheses. This indicates that our method converges faster with the help of DPVL compared to the original PVNet. Therefore, our network achieves convergence within 100 while PVNet requires 200 epochs. 
Figure.~\ref{fig:training} also implies that the distribution of hypotheses produced by our method is more concentrated since the mean distance between proxy hypotheses and keypoints is smaller. This enables our method to reach consensus more easily when voting keypoints. 

\vspace{-0.5em}
\section{Conclusion}
\vspace{-0.5em}
In this paper, we proposed a novel differentiable proxy voting loss (DPVL) to achieve accurate vector-field estimation by mimicking the hypothesis voting procedure in the testing phase. Our DPVL takes the distances between pixels and the keypoints into account and enforces pixels away from a keypoint to be intolerant to inaccurate direction vector estimation. In this manner, our method achieves more accurate vector fields, thus leading to better pose estimation. Moreover, DPVL is able to speed up the convergence of our network in training. Thus, our method requires fewer iterations in training but achieves better testing performance compared to our baseline method PVNet.
Extensive experiments on two standard pose estimation datasets demonstrate the superiority of our proposed method. 

\newpage
\section{More qualitative results}
Below, we provide more visual results of our method.

\begin{figure*}[!ht]
\centering
{\includegraphics[width=1\linewidth]{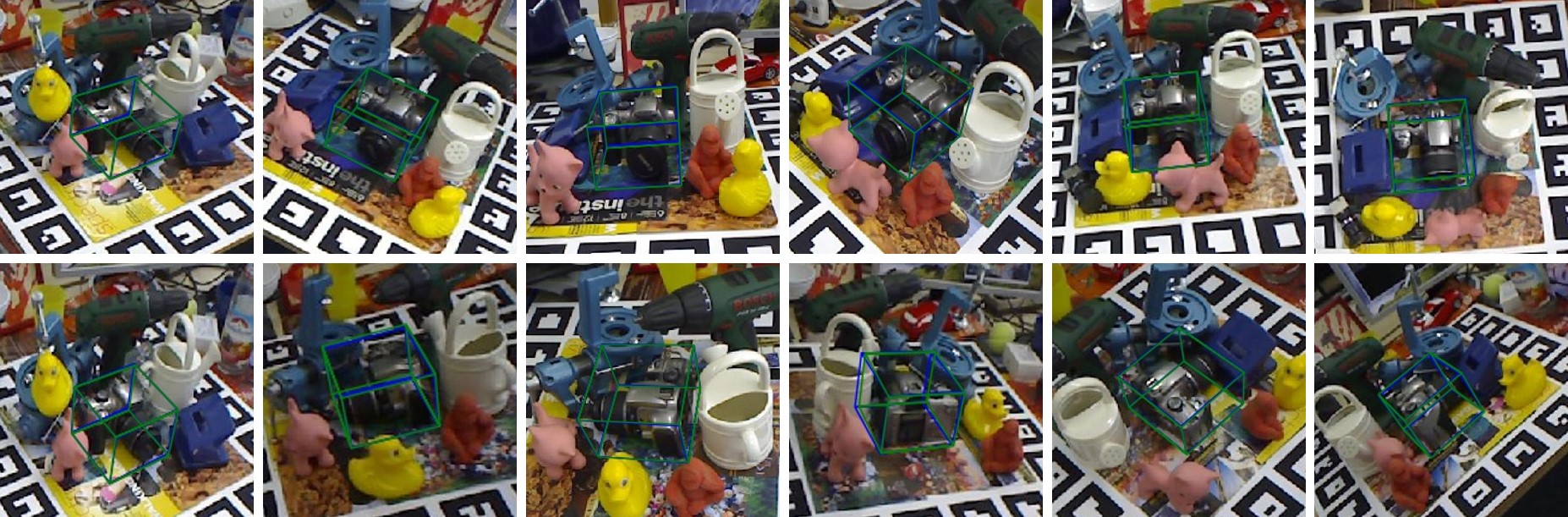}}\vspace{0.8em}
{\includegraphics[width=1\linewidth]{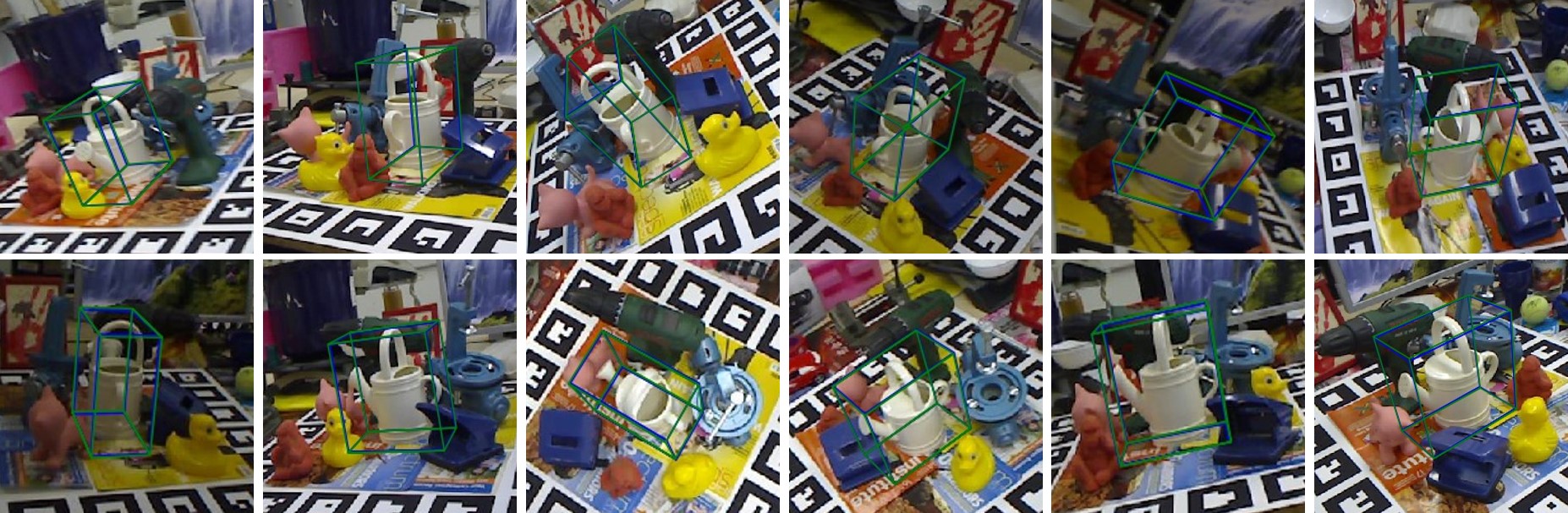}}\vspace{0.8em}
{\includegraphics[width=1\linewidth]{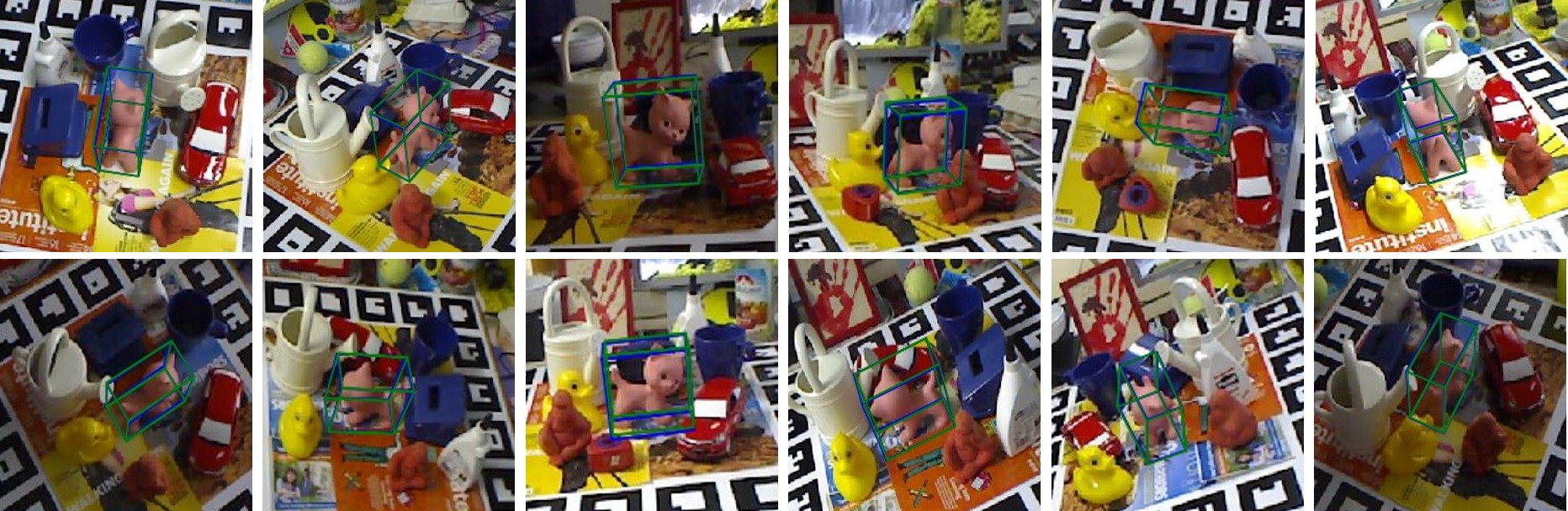}}\vspace{0.8em}
\vspace{-1em}
\caption{Visualization of qualitative results on the {\bf LINEMOD} dataset. Green 3D bounding-boxes represent the ground-truth poses and the blue ones indicate our predictions.}
\end{figure*}

\begin{figure*}[!ht]
\centering
{\includegraphics[width=1\linewidth]{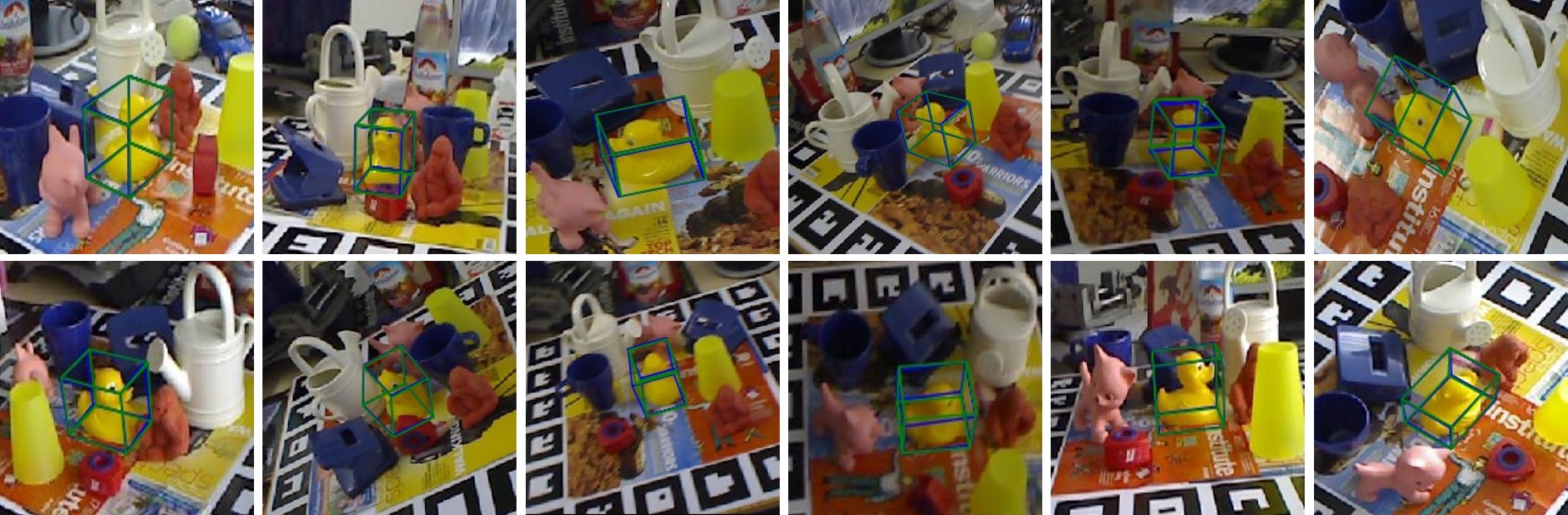}}\vspace{0.8em}
{\includegraphics[width=1\linewidth]{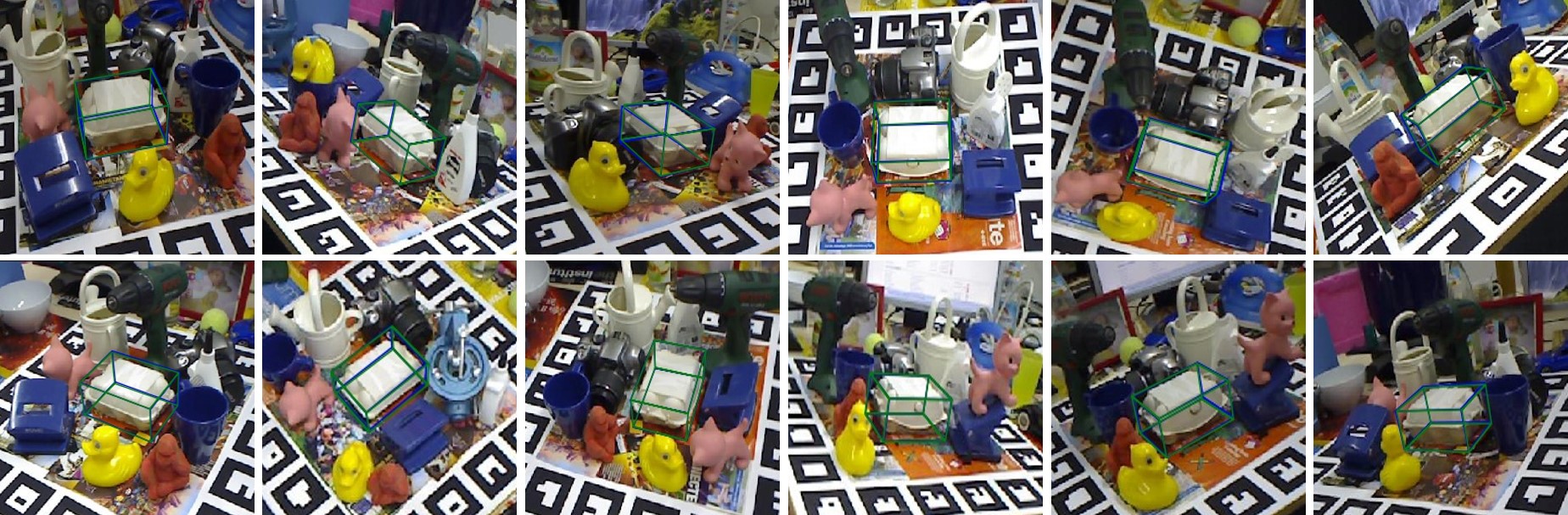}}\vspace{0.8em}
{\includegraphics[width=1\linewidth]{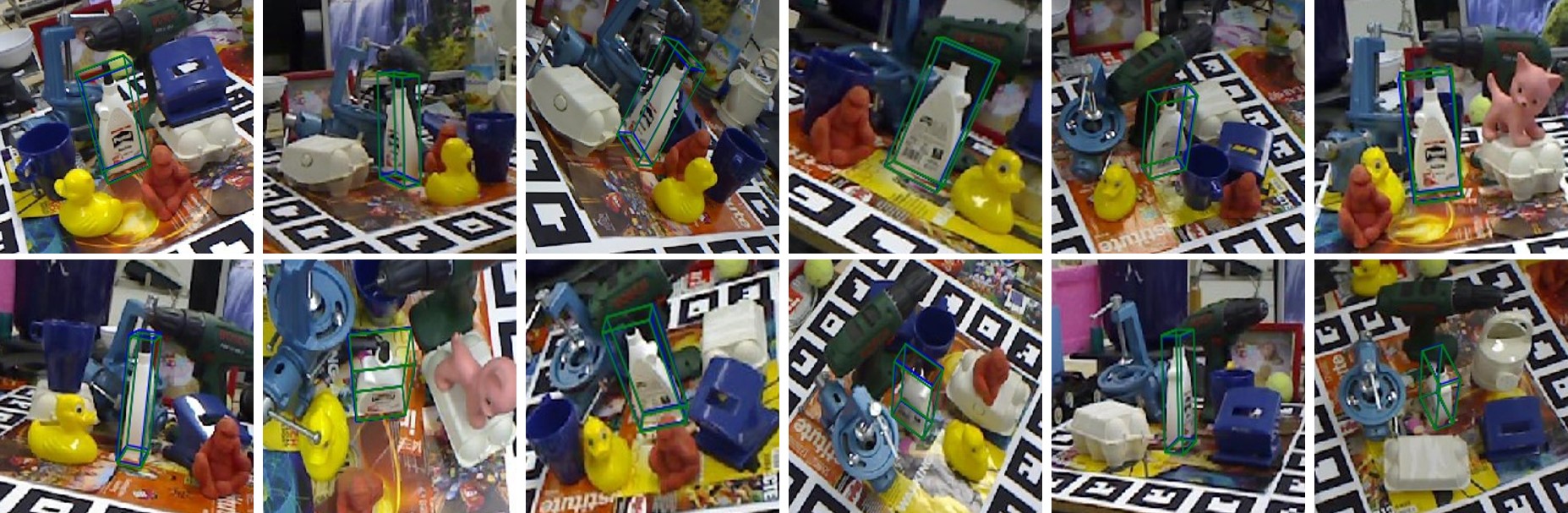}}\vspace{0.8em}
\vspace{-1em}
\caption{Visualization of qualitative results on the {\bf LINEMOD} dataset. Green 3D bounding-boxes represent the ground-truth poses and the blue ones indicate our predictions.}
\end{figure*}

\begin{figure*}[!ht]
\centering
{\includegraphics[width=1\linewidth]{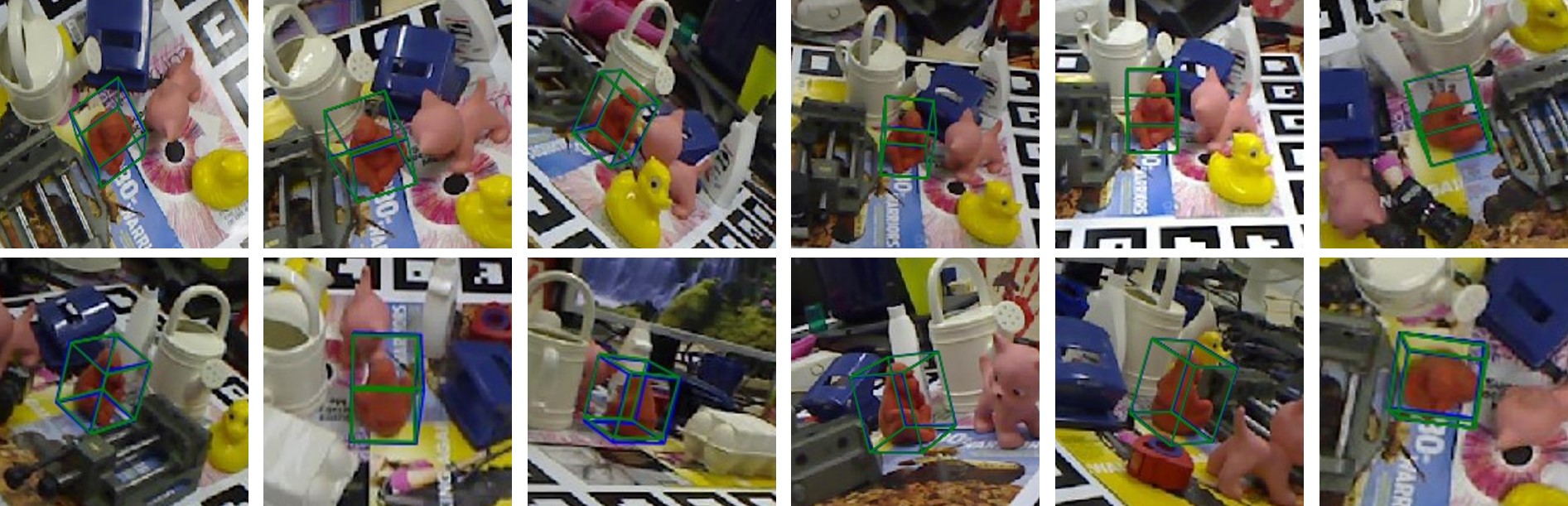}}\vspace{0.8em}
{\includegraphics[width=1\linewidth]{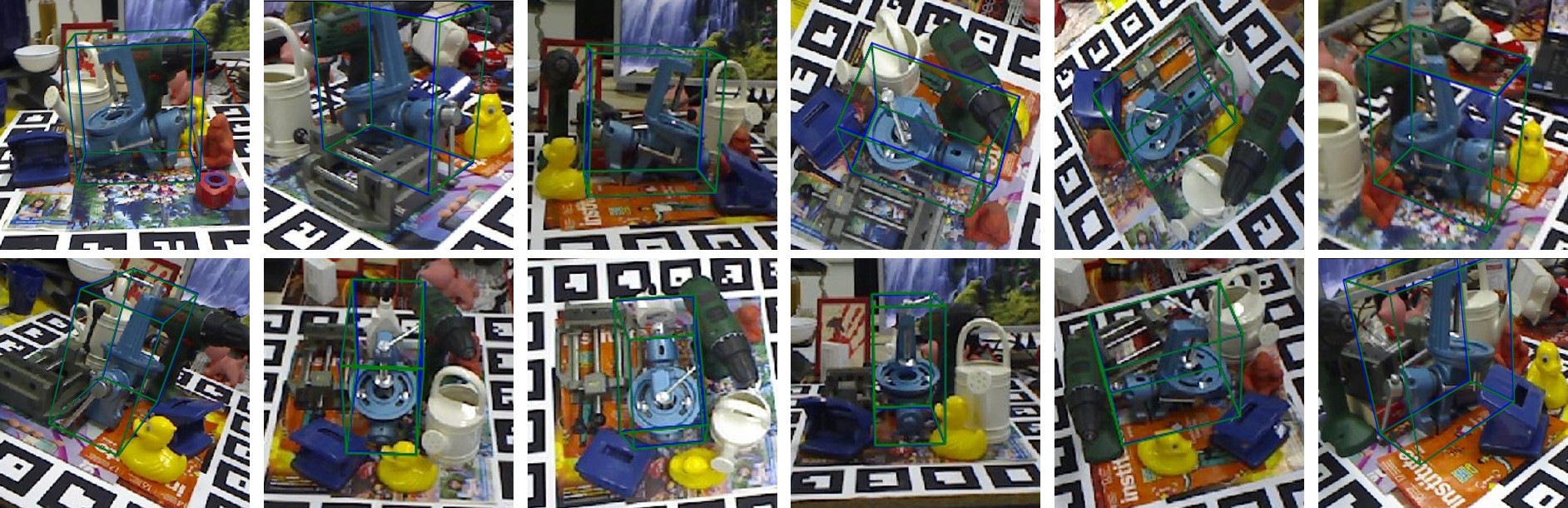}}\vspace{0.8em}
\vspace{-1em}
\caption{Visualization of qualitative results on the {\bf LINEMOD} dataset. Green 3D bounding-boxes represent the ground-truth poses and the blue ones indicate our predictions.}
\end{figure*}

\begin{figure*}[!ht]
\centering
{\includegraphics[width=1\linewidth]{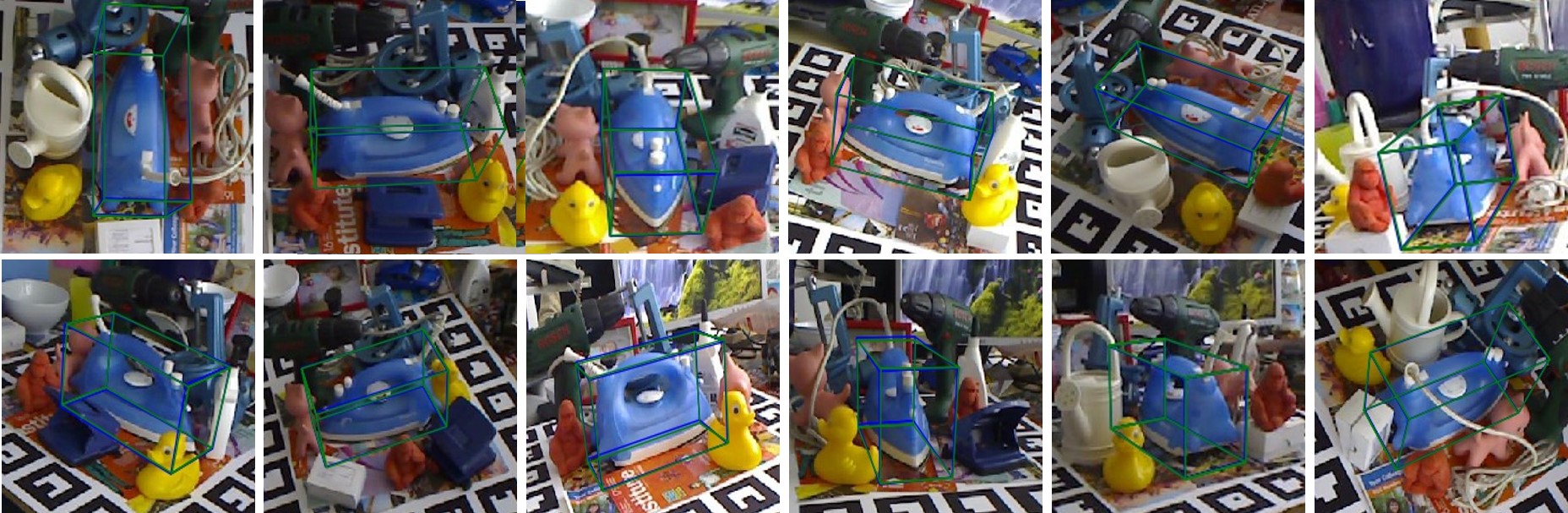}}\vspace{0.8em}
{\includegraphics[width=1\linewidth]{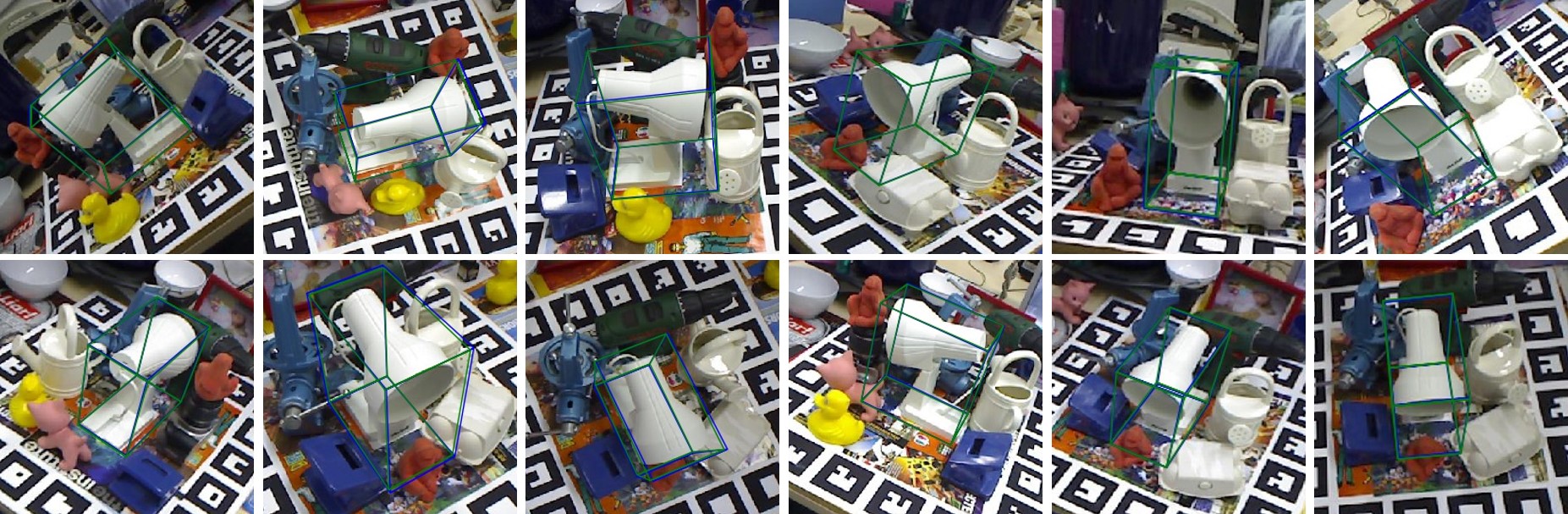}}\vspace{0.8em}
{\includegraphics[width=1\linewidth]{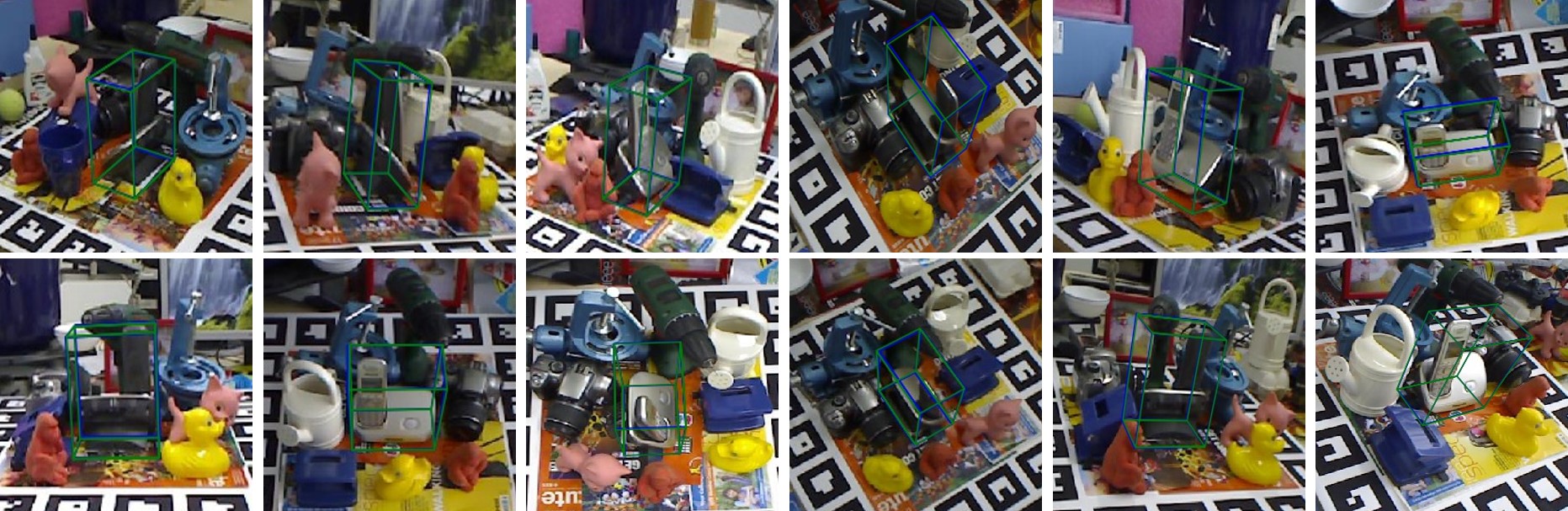}}
\vspace{-1em}
\caption{Visualization of qualitative results on the {\bf LINEMOD} dataset. Green 3D bounding-boxes represent the ground-truth poses and the blue ones indicate our predictions.}
\end{figure*}

\begin{figure*}[!ht]
\centering
{\includegraphics[width=1\linewidth]{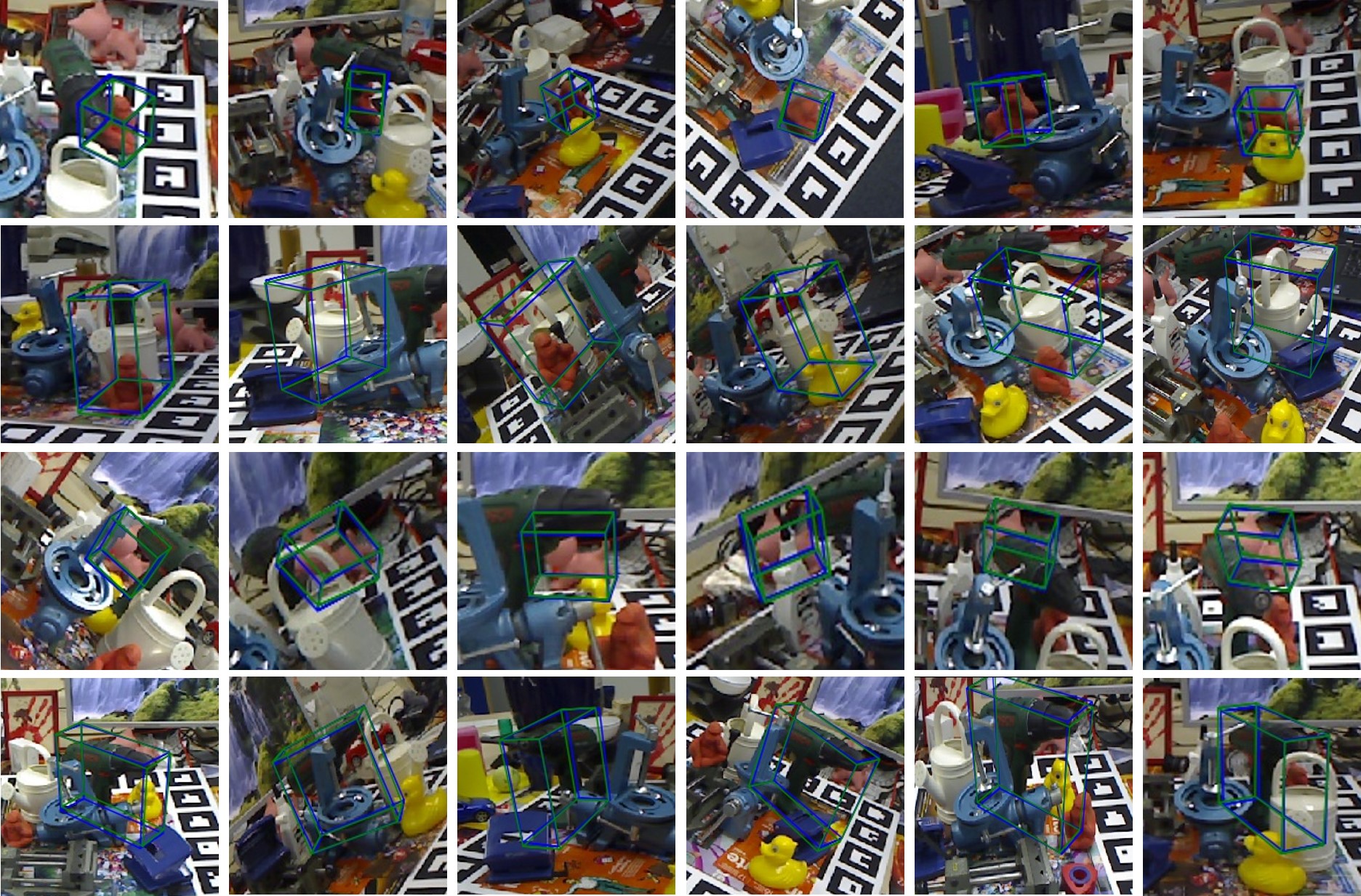}}
\vspace{-1em}
\caption{Visualization of qualitative results on the {\bf Occlusion LINEMOD} dataset. Green 3D bounding-boxes represent the ground-truth poses and the blue ones indicate our predictions.}
\end{figure*}

\begin{figure*}[!ht]
\centering
{\includegraphics[width=1\linewidth]{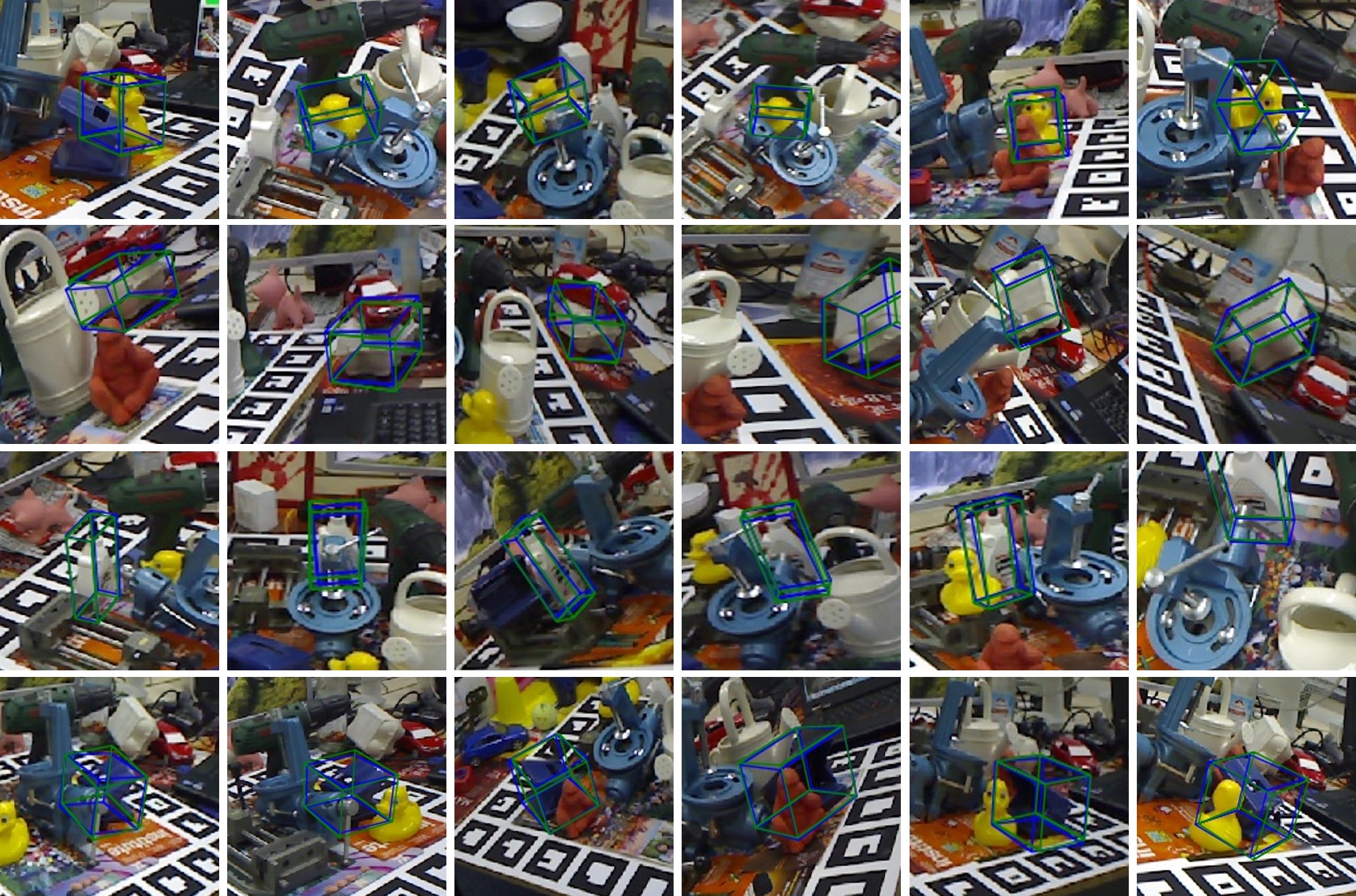}}
\vspace{-1em}
\caption{Visualization of qualitative results on the {\bf Occlusion LINEMOD} dataset. Green 3D bounding-boxes represent the ground-truth poses and the blue ones indicate our predictions.}
\end{figure*}


\begin{thebibliography}{33}
\providecommand{\natexlab}[1]{#1}
\providecommand{\url}[1]{\texttt{#1}}
\expandafter\ifx\csname urlstyle\endcsname\relax
  \providecommand{\doi}[1]{doi: #1}\else
  \providecommand{\doi}{doi: \begingroup \urlstyle{rm}\Url}\fi

\bibitem[Besl and McKay(1992)]{besl1992method}
Paul~J Besl and Neil~D McKay.
\newblock Method for registration of 3-d shapes.
\newblock In \emph{Sensor fusion IV: control paradigms and data structures},
  volume 1611, pages 586--606. International Society for Optics and Photonics,
  1992.

\bibitem[Brachmann et~al.(2014)Brachmann, Krull, Michel, Gumhold, Shotton, and
  Rother]{brachmann2014learning}
Eric Brachmann, Alexander Krull, Frank Michel, Stefan Gumhold, Jamie Shotton,
  and Carsten Rother.
\newblock Learning 6d object pose estimation using 3d object coordinates.
\newblock In \emph{European Conference on Computer Vision (ECCV)}, pages
  536--551. Springer, 2014.

\bibitem[Chen et~al.(2017)Chen, Badrinarayanan, Lee, and
  Rabinovich]{chen2017gradnorm}
Zhao Chen, Vijay Badrinarayanan, Chen-Yu Lee, and Andrew Rabinovich.
\newblock Gradnorm: Gradient normalization for adaptive loss balancing in deep
  multitask networks.
\newblock \emph{arXiv preprint arXiv:1711.02257}, 2017.

\bibitem[Fischler and Bolles(1981)]{ransac}
Martin~A. Fischler and Robert~C. Bolles.
\newblock Random sample consensus: A paradigm for model fitting with
  applications to image analysis and automated cartography.
\newblock \emph{Commun. ACM}, 1981.

\bibitem[Girshick(2015)]{girshick2015fast}
Ross Girshick.
\newblock Fast r-cnn.
\newblock In \emph{Proceedings of the IEEE International Conference on Computer
  Vision (ICCV)}, pages 1440--1448, 2015.

\bibitem[Gu and Ren(2010)]{gu2010discriminative}
Chunhui Gu and Xiaofeng Ren.
\newblock Discriminative mixture-of-templates for viewpoint classification.
\newblock In \emph{European Conference on Computer Vision (ECCV)}, pages
  408--421. Springer, 2010.

\bibitem[He et~al.(2016)He, Zhang, Ren, and Sun]{he2016deep}
Kaiming He, Xiangyu Zhang, Shaoqing Ren, and Jian Sun.
\newblock Deep residual learning for image recognition.
\newblock In \emph{Proceedings of the IEEE conference on Computer Vision and
  Pattern Recognition (CVPR)}, pages 770--778, 2016.

\bibitem[Hinterstoisser et~al.(2011{\natexlab{a}})Hinterstoisser, Cagniart,
  Ilic, Sturm, Navab, Fua, and Lepetit]{hinterstoisser2011gradient}
Stefan Hinterstoisser, Cedric Cagniart, Slobodan Ilic, Peter Sturm, Nassir
  Navab, Pascal Fua, and Vincent Lepetit.
\newblock Gradient response maps for real-time detection of textureless
  objects.
\newblock \emph{IEEE Transactions on Pattern Analysis and Machine
  Intelligence}, 34\penalty0 (5):\penalty0 876--888, 2011{\natexlab{a}}.

\bibitem[Hinterstoisser et~al.(2011{\natexlab{b}})Hinterstoisser, Holzer,
  Cagniart, Ilic, Konolige, Navab, and Lepetit]{hinterstoisser2011multimodal}
Stefan Hinterstoisser, Stefan Holzer, Cedric Cagniart, Slobodan Ilic, Kurt
  Konolige, Nassir Navab, and Vincent Lepetit.
\newblock Multimodal templates for real-time detection of texture-less objects
  in heavily cluttered scenes.
\newblock In \emph{Proceedings of the IEEE International Conference on Computer
  Vision (ICCV)}, pages 858--865. IEEE, 2011{\natexlab{b}}.

\bibitem[Hinterstoisser et~al.(2012)Hinterstoisser, Lepetit, Ilic, Holzer,
  Bradski, Konolige, and Navab]{hinterstoisser2012model}
Stefan Hinterstoisser, Vincent Lepetit, Slobodan Ilic, Stefan Holzer, Gary
  Bradski, Kurt Konolige, and Nassir Navab.
\newblock Model based training, detection and pose estimation of texture-less
  3d objects in heavily cluttered scenes.
\newblock In \emph{Asian Conference on Computer Vision (ACCV)}, pages 548--562.
  Springer, 2012.

\bibitem[Hu et~al.(2019{\natexlab{a}})Hu, Fua, Wang, and
  Salzmann]{hu2019single}
Yinlin Hu, Pascal Fua, Wei Wang, and Mathieu Salzmann.
\newblock Single-stage 6d object pose estimation.
\newblock \emph{arXiv preprint arXiv:1911.08324}, 2019{\natexlab{a}}.

\bibitem[Hu et~al.(2019{\natexlab{b}})Hu, Hugonot, Fua, and
  Salzmann]{hu2019segmentation}
Yinlin Hu, Joachim Hugonot, Pascal Fua, and Mathieu Salzmann.
\newblock Segmentation-driven 6d object pose estimation.
\newblock In \emph{Proceedings of the IEEE Conference on Computer Vision and
  Pattern Recognition (CVPR)}, pages 3385--3394, 2019{\natexlab{b}}.

\bibitem[Kehl et~al.(2017)Kehl, Manhardt, Tombari, Ilic, and
  Navab]{kehl2017ssd}
Wadim Kehl, Fabian Manhardt, Federico Tombari, Slobodan Ilic, and Nassir Navab.
\newblock Ssd-6d: Making rgb-based 3d detection and 6d pose estimation great
  again.
\newblock In \emph{Proceedings of the IEEE International Conference on Computer
  Vision (ICCV)}, pages 1521--1529, 2017.

\bibitem[Kiru et~al.(2019)Kiru, Timothy, and Markus]{Park2019Pix2Pose}
Park Kiru, Patten Timothy, and Vincze Markus.
\newblock Pix2pose: Pixel-wise coordinate regression of objects for 6d pose
  estimation.
\newblock In \emph{Proceedings of the IEEE International Conference on Computer
  Vision (ICCV)}, 2019.

\bibitem[Lepetit et~al.(2009)Lepetit, Moreno-Noguer, and Fua]{lepetit2009epnp}
Vincent Lepetit, Francesc Moreno-Noguer, and Pascal Fua.
\newblock Epnp: An accurate o (n) solution to the pnp problem.
\newblock \emph{International Journal of Computer Vision}, 81\penalty0
  (2):\penalty0 155, 2009.

\bibitem[Li et~al.(2018)Li, Wang, Ji, Xiang, and Fox]{li2018deepim}
Yi~Li, Gu~Wang, Xiangyang Ji, Yu~Xiang, and Dieter Fox.
\newblock Deepim: Deep iterative matching for 6d pose estimation.
\newblock In \emph{European Conference on Computer Vision (ECCV)}, pages
  683--698, 2018.

\bibitem[Liu et~al.(2010)Liu, Tuzel, Veeraraghavan, and Chellappa]{liu2010fast}
Ming-Yu Liu, Oncel Tuzel, Ashok Veeraraghavan, and Rama Chellappa.
\newblock Fast directional chamfer matching.
\newblock In \emph{Proceedings of the IEEE Conference on Computer Vision and
  Pattern Recognition (CVPR)}, pages 1696--1703. IEEE, 2010.

\bibitem[Liu et~al.(2016)Liu, Anguelov, Erhan, Szegedy, Reed, Fu, and
  Berg]{liu2016ssd}
Wei Liu, Dragomir Anguelov, Dumitru Erhan, Christian Szegedy, Scott Reed,
  Cheng-Yang Fu, and Alexander~C Berg.
\newblock Ssd: Single shot multibox detector.
\newblock In \emph{European conference on computer vision (ECCV)}, pages
  21--37. Springer, 2016.

\bibitem[Lowe(2004)]{lowe2004distinctive}
David~G Lowe.
\newblock Distinctive image features from scale-invariant keypoints.
\newblock \emph{International journal of computer vision}, 60\penalty0
  (2):\penalty0 91--110, 2004.

\bibitem[Oberweger et~al.(2018)Oberweger, Rad, and
  Lepetit]{oberweger2018making}
Markus Oberweger, Mahdi Rad, and Vincent Lepetit.
\newblock Making deep heatmaps robust to partial occlusions for 3d object pose
  estimation.
\newblock In \emph{European Conference on Computer Vision (ECCV)}, pages
  119--134, 2018.

\bibitem[Peng et~al.(2019)Peng, Liu, Huang, Zhou, and Bao]{peng2019pvnet}
Sida Peng, Yuan Liu, Qixing Huang, Xiaowei Zhou, and Hujun Bao.
\newblock Pvnet: Pixel-wise voting network for 6dof pose estimation.
\newblock In \emph{Proceedings of the IEEE Conference on Computer Vision and
  Pattern Recognition (CVPR)}, pages 4561--4570, 2019.

\bibitem[Rad and Lepetit(2017)]{rad2017bb8}
Mahdi Rad and Vincent Lepetit.
\newblock Bb8: A scalable, accurate, robust to partial occlusion method for
  predicting the 3d poses of challenging objects without using depth.
\newblock In \emph{Proceedings of the IEEE International Conference on Computer
  Vision (CVPR)}, pages 3828--3836, 2017.

\bibitem[Ren et~al.(2015)Ren, He, Girshick, and Sun]{ren2015faster}
Shaoqing Ren, Kaiming He, Ross Girshick, and Jian Sun.
\newblock Faster r-cnn: Towards real-time object detection with region proposal
  networks.
\newblock In \emph{Advances in Neural Information Processing Systems}, pages
  91--99, 2015.

\bibitem[Rios-Cabrera and Tuytelaars(2013)]{rios2013discriminatively}
Reyes Rios-Cabrera and Tinne Tuytelaars.
\newblock Discriminatively trained templates for 3d object detection: A real
  time scalable approach.
\newblock In \emph{Proceedings of the IEEE International Conference on Computer
  Vision (ICCV)}, pages 2048--2055, 2013.

\bibitem[Song et~al.(2020)Song, Song, and Huang]{song2020hybridpose}
Chen Song, Jiaru Song, and Qixing Huang.
\newblock Hybridpose: 6d object pose estimation under hybrid representations.
\newblock \emph{arXiv preprint arXiv:2001.01869}, 2020.

\bibitem[Su et~al.(2015)Su, Qi, Li, and Guibas]{su2015render}
Hao Su, Charles~R Qi, Yangyan Li, and Leonidas~J Guibas.
\newblock Render for cnn: Viewpoint estimation in images using cnns trained
  with rendered 3d model views.
\newblock In \emph{Proceedings of the IEEE International Conference on Computer
  Vision (ICCV)}, pages 2686--2694, 2015.

\bibitem[Tekin et~al.(2018)Tekin, Sinha, and Fua]{tekin2018real}
Bugra Tekin, Sudipta~N Sinha, and Pascal Fua.
\newblock Real-time seamless single shot 6d object pose prediction.
\newblock In \emph{Proceedings of the IEEE Conference on Computer Vision and
  Pattern Recognition (CVPR)}, pages 292--301, 2018.

\bibitem[Tulsiani and Malik(2015)]{tulsiani2015viewpoints}
Shubham Tulsiani and Jitendra Malik.
\newblock Viewpoints and keypoints.
\newblock In \emph{Proceedings of the IEEE Conference on Computer Vision and
  Pattern Recognition (CVPR)}, pages 1510--1519, 2015.

\bibitem[Wang et~al.(2019)Wang, Sridhar, Huang, Valentin, Song, and
  Guibas]{wang2019normalized}
He~Wang, Srinath Sridhar, Jingwei Huang, Julien Valentin, Shuran Song, and
  Leonidas~J Guibas.
\newblock Normalized object coordinate space for category-level 6d object pose
  and size estimation.
\newblock In \emph{Proceedings of the IEEE Conference on Computer Vision and
  Pattern Recognition (CVPR)}, pages 2642--2651, 2019.

\bibitem[Xiang et~al.(2017)Xiang, Schmidt, Narayanan, and
  Fox]{xiang2017posecnn}
Yu~Xiang, Tanner Schmidt, Venkatraman Narayanan, and Dieter Fox.
\newblock Posecnn: A convolutional neural network for 6d object pose estimation
  in cluttered scenes.
\newblock \emph{Robotics: Science and Systems}, 2017.

\bibitem[Zakharov et~al.(2019)Zakharov, Shugurov, and Ilic]{zakharovdpod}
Sergey Zakharov, Ivan Shugurov, and Slobodan Ilic.
\newblock Dpod: 6d pose object detector and refiner.
\newblock In \emph{Proceedings of the IEEE International Conference on Computer
  Vision (ICCV)}, 2019.

\bibitem[Zhigang et~al.(2019)Zhigang, Gu, and Xiangyang]{li2019CDPN}
Li~Zhigang, Wang Gu, and Ji~Xiangyang.
\newblock Cdpn: Coordinates-based disentangled pose network for real-time
  rgb-based 6-dof object pose estimation.
\newblock In \emph{Proceedings of the IEEE International Conference on Computer
  Vision (ICCV)}, 2019.

\bibitem[Zhu et~al.(2014)Zhu, Derpanis, Yang, Brahmbhatt, Zhang, Phillips,
  Lecce, and Daniilidis]{zhu2014single}
Menglong Zhu, Konstantinos~G Derpanis, Yinfei Yang, Samarth Brahmbhatt, Mabel
  Zhang, Cody Phillips, Matthieu Lecce, and Kostas Daniilidis.
\newblock Single image 3d object detection and pose estimation for grasping.
\newblock In \emph{IEEE International Conference on Robotics and Automation
  (ICRA)}, pages 3936--3943. IEEE, 2014.

\end{thebibliography}

\end{document}